*Chapter 3*

# A Virtual-Based Haptic Endoscopic Sinus Surgery (ESS) Training System: from Development to Validation

Soroush Sadeghnejad[*], Mojtaba Esfandiari, Farshad Khadivar

*Bioinspired System Design Laboratory, Biomedical Engineering Department, AmirKabir University of Technology (Tehran Polytechnic), No. 350, Hafez Ave, Valiasr Square, Tehran, Iran 1591634311, I.R. IRAN. Emails: s.sadeghnejad@aut.ac.ir, fkhadivar20@gmail.com, Mojtaba.esfandiari.92@gmail.com,*
[*]*Corresponding author*

**Abstract** – Simulated training platforms offer a suitable avenue for surgical students and professionals to build and improve upon their skills, without the hassle of traditional training methods. To enhance the degree of realistic interaction paradigms of training simulators, great work has been done to both model simulated anatomy in more realistic fashion, as well as providing appropriate haptic feedback to the trainee. As such, this chapter seeks to discuss the ongoing research being conducted on haptic feedback incorporated simulators specifically for Endoscopic Sinus Surgery (ESS). This chapter offers a brief comparative analysis of some EES simulators, in addition to a deeper quantitative and qualitative look into our approach to designing and prototyping a complete virtual-based haptic EES training platform.

## 3.1 INTRODUCTION

With the integration of robotic systems in surgery, the adaptability and success rate of surgery has improved noticeably, allowing for surgeons to automate repetitive tasks, reduce the manpower in the OR, as well as reduce the risk posed to the patient by directly alleviating surgeon fatigue (Taylor *et al* 1995) (Casals 1998) (Michel 2021). Another critical factor that is addressed through the introduction of robotics in surgery is the high level of skill that is demanded from the surgeon; highly delicate surgeries require years of training, in addition to an exceptional understanding of the human anatomy. ESS, characteristically a minimally invasive Endoscopic Sinus Surgery, is one of such surgeries (Fried *et al* 2005) (Zhao *et al* 2021) (Lourijsen *et al* 2022). Given the tight spatial and visual constraints, the increased complexity of the procedure demands the ability to navigate around intraoperative issues such as visual perception, anatomy recognition, and nonhomogeneous anatomical makeup, not to mention the real-time identification of presence of critical regions like brain tissue, carotid artery, optic nerve, and other intracranial structures (Fried *et al* 2004). Thus, the importance of extensive practice and training is undoubtedly high for increasing the success rate for such a surgery.

As such, the need for accurate and adaptable training systems is imperative, in the light of limited opportunities to observe or practice live and/or on cadavers, not to mention the abundance of ethical, monetary, and safety considerations associated with the traditional training methodologies. The suggested training simulation platforms afford repeatable, closely monitored, and adaptable training opportunities, with the added benefit of standardizing a benchmark of acceptable surgical skill that can be applied across any participating group of trainees. The level of training, however, is proportional to how closely the simulated training system matches a real surgical procedure; all the way from the preoperative protocol to the intraoperative tissue dynamics to the feel of the tool-anatomy interactions (Piromchai 2014) (Samur *et al* 2007) (Chanthasopeephan *et al* 2003).

The development of ESS simulators provided a more efficient training method, compared to the conventional methods, and enabled the trainers and trainees to safely and frequently practice surgical operations on a standard simulated environment which results in reduced training time, costs, risks, to name but a few. In the early stages, most of the VR-based surgical training simulators were mainly based on learning through imitation of simulated surgical operations but lacked surgical skill evaluation and realistic sense of haptic feedback during tool-tissue interaction (Perez-Gutierrez *et al* 2010). Survey results demonstrate that training surgeons and residents with surgical VR-based simulators minimizes the risk on patients during real surgeries (Zhao *et al* 2011). However, new technologies in robotic actuation and computational sensing methods contributed to the advancement of a wide variety of surgical haptic training simulators with different types of feedback modes. In these systems, the human operator interacts with a simulated environment using a haptic interface, usually equipped with some sort of force, vibration, sound, etc. feedback. The simulated environment can be physical, virtual, or physical-virtual (hybrid) (Lalitharatne *et al* 2020)

## 3.2 HISTORY AND STATE OF THE ART

The first virtual reality (VR) sinus surgery simulator was developed in late 2000. The ES3, a non-commercialized prototype developed by Lockheed Martin, Akron, OH (Wiet *et al* 2008), offered a series of increasingly challenging simulation exercises in which trainees could master their skills by focusing on the learning courses of a complete sinus surgery procedure. Force feedback and the computer graphics were integral in creating a virtual surgical environment, and training on this platform translated into performance improvement in the OR. The system, which comprises of a simulation platform, a haptic controller, a voice-recognition controlled platform (to control the simulator), and a dummy setup for human interaction, further records, analyses, and report performance metrics in real time (Fried *et al* 2007).

The Dextroscope endoscopic sinus simulator (Caversaccio *et al* 2003) is one of the other non-commercialized prototypes, developed for supporting the training purposes. With two hand-held tools (one for controlling precision and one for volume manipulation) and stereoscopic goggles, the user can interact with both a panel, and segmented virtual models of the endonasal region, represented as 3D volumetric data. However, this system did not translate into improved performance in the OR, and the lack of force feedback was also an issue.

To fill the void of acceptability and validity in ESS training, the VOXEL-MAN sinus surgery simulator was developed in 2010 (Tolsdorff *al* 2010). The system is equipped with a 3D model of sinus and nasal cavity environment based on a high-resolution computed tomography, which can be manipulated with virtual surgical tools, controlled with a low-cost haptic device, yet providing haptic rendering and tissue removal visualization for training purposes (Piromchai 2004).

In addition to actual setups, there have been attempts to improving the paradigm of tissue and/or anatomical modelling for these training simulators. In 2009, Parikh et al., proposed a new method for an automatic construction of the patient's 3D model anatomy for virtual surgical environment from preoperative CT images.

Subsequently in 2010, Perez-Gutierrez *et al.*, proposed a 4-DoFs endoscopic endonasal simulator prototype in which they simulated a rigid endoscope movement, a simplified nasal tissue collision, and a contact force model for haptic feedback. Users can simulate movements such as the insertion of the endoscope into nasal cavities and receive haptic feedback modelled by a damped mass-spring model (Perez-Gutierrez *et al* 2010).

The National Research Council of Canada developed a VR simulator, NeuroTouch (Delorme *et al* 2012), for cranial micro neurosurgery training with haptic and graphics feedbacks. The mechanical behaviour of tissues is modelled as that of viscoelastic solids; the elastic property is modelled as a hyperplastic solid using a generalized Rivlin constitutive model, while the viscous property is modelled by a quasilinear viscoelastic constitutive model. This commercialized simulator is used in 7 teaching hospitals across Canada to investigate the system performance and validate the system behaviour (Varshney *et al* 2014).

Another proposed product, NeuroTouch-Endo VR, is a VR simulator for training neurosurgeons, and is built upon the NeuroTouch system. This system also provides haptic feedback and can be used for teaching endonasal endoscopic transsphenoidal surgery, given its ability to record, analyse, and report performance metrics (Rosseau *et al* 2013).

A VR simulator of sinus surgery with haptic feedback, the McGill simulator for endoscopic sinus surgery (MSESS), was developed by collaboration between McGill University, the Montreal Neurologic Institute Simulation Lab, and the National Research Council of Canada. The objectives of this study were to show the acceptability, perceived realism of the developed system among technical users, and to evaluate the training progress via some performance metrics (Varshney *et al* 2016).

In 2018, Barber *et al* developed a modular VR teaching tool for surgical training purposes. They utilized three-dimensional models to represent the critical anatomic structures. A 3D-printed skull model and a virtual endoscope were designed to develop a low-cost educational device (Barber *et al* 2018).

Kim *et al* proposed a VR haptic simulator using patient-specific 3D–printed external nostril and an exchangeable caudal septum model that facilitates real surgical simulation for training of endoscopic sinus surgery (Kim *et al* 2020).

Computer tomography (CT) images are used to generate a graphical model for the virtual environment. The training outcomes of recruited subjects were evaluated by defining some vivid parameters (Pößneck *et al* 2022).

Table 3.1 reveals a summary of the developed VR endoscopic sinus and skull base surgery systems' specifications.

As such, many other training simulators have been developed and launched over the years, each trying to address the issue of providing realistic training simulation for sinus surgery. Some research groups aim to achieve realistic tissue and anatomical modelling, in a bid to offer genuine haptic feedback during the training session, while some groups focus on enhancing the control schemes implemented in their systems, trying to overcome the paradigm of stability and haptic transparency. This chapter provides a comprehensive look into the various approaches in developing an ESS simulator that offers trainees a viable intraoperative experience through different models for haptic feedback of tooltip and anatomy interaction, in addition to offering an opportunity for skill enhancement of the participants via a variety of tests.

**Table 3.1**. Virtual reality simulators developed for endoscopic sinus surgery

| Simulators | Characteristics | Real-based tissue mechanical model | Force-feedback control | Haptic feedback |
|---|---|---|---|---|
| ES3 | Stereoscopic display, haptic devices, voice recognition | No | No | Yes |
| **Dextroscope** | Stereoscopic display, haptic devices, real-based 3D model environment | No | No | No |
| **VOXEL-MAN Sinus Surgery** | Stereoscopic display, haptic devices | No | No | Yes |
| **Stanford** | real-based 3D visuo-haptic models, haptic devices | No | No | Yes |
| **Perez-Gutierrez simulator** | user-endoscope-tissue model, haptic rendering | No | No | Yes |
| **NeuroTouch** | stereovision system, haptic tool manipulators, high-end computers | Yes | No | Yes |
| **NeuroTouch-Endo VR** | stereovision system, haptic tool manipulators, high-end computers | No | No | Yes |
| **McGill simulator** | stereovision system, haptic tool manipulators, high-end computers | No | No | Yes |
| **Barber et al simulator** | Stereoscopic display, real-based 3D model environment | No | No | No |
| **Kim et al simulator** | haptic devices, patient specific 3D model environment | Yes | No | Yes |
| **Pößneck et al simulator** | haptic devices, 3D model environment | Yes | No | Yes |

We will build upon pre-existing and established projects revolving around the fundamentals of endoscopic sinus surgery as standardized training modules for technical skills and provide a roadmap for other academics to produce a realistic ESS training simulator (Sadeghnejad *et al* 2016) (Khadivar *et al* 2017) (Ebrahimi *et al* 2016). We will also discuss the broad results, offering the reader an insight into how to develop, as well as validate, a well-rounded ESS training simulator (Sadeghnejad *et al* 2019).

## 3.3 DEVELOPMENT OF AN ENDOSCOPIC SINUS TRAINING SYSTEM

The development of an appropriate ESS training simulator is a multifaceted process, starting from identifying the key skills that the trainees will be training and be assessed on (Figure 3.1). Following this, there needs to be a careful selection of the hardware necessary to deliver the target feature, including but not limited to a virtual reality platform and supporting equipment, haptic feedback platform, computers with the target specifications, and so forth. Another critical element is the methodology that is to be used to model the anatomy/tissue of the endonasal region. The importance of this factor is paramount given the selected mathematical model drives the design of virtual interaction force feedback. Furthermore, there also needs to be a specific control structure to manage the force feedback loop. Moreover, there needs to be a comprehensive approach to testing the proposed system, which is essential for evaluating the effectiveness of the system.

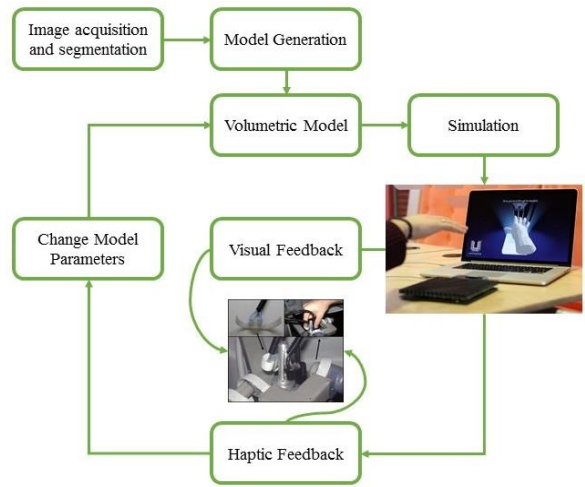

**Figure 3.1**. Components of an endoscopic sinus surgery training system development

### 3.3.1 Core technical skills identification

The development of an Endoscopic Sinus Surgery training simulator in the Djavad Mowafaghian Research Centre of Intelligent Neurorehabilitation Technologies (DMRCINT), Mechanical Engineering Department, Sharif University of Technology, Tehran, Iran, was an effort of a team to focus on the training of endoscopic sinus surgery skills and prioritizing performance objectives that involved hands-on techniques. After consulting about broad categories of identified skills in the ESS, we highlighted the force feedback hands-on techniques which would enable the trainees to identify key haptic feedbacks revolving around the interaction of an operative/surgical tool with key vital structures as an orbits and carotid artery present through the intricate anatomy of the paranasal sinuses (Figure 3.2).

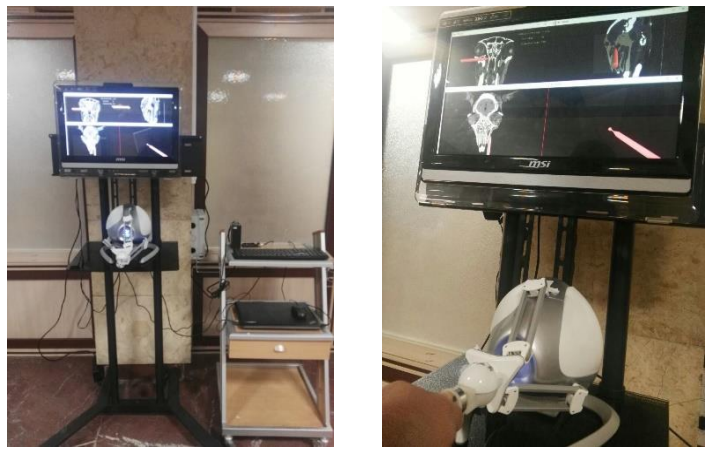

**Figure 3.2**. An endoscopic sinus surgery training system, equipped with a haptic interface, a virtual complete model, and the endoscopic views

Furthermore, it is naturally critical that trainees are given access to the training simulator for an appropriate amount of time to allow for an acceptable level of skill acquisition and become comfortable with the user and haptic interfaces as they pertain to the simulated orbital floor removal in an actual ESS operation. This was achieved by having the participants partake in three tasks: pre-experiment learning, training, and evaluation tasks.

### 3.3.2 Hardware

The VR simulator consists of a computer, virtual graphics rendering system (for VR application), a haptic user interface, force sensor and data acquisition instruments, and a video monitoring system (Figure 3.3). The training system can allow the user to manipulate tools and instruments commonly used during endonasal transsphenoidal surgery, e.g., curettes. The tool-tissue interaction force in the virtual environment is replicated on the user hand through the haptic interface.

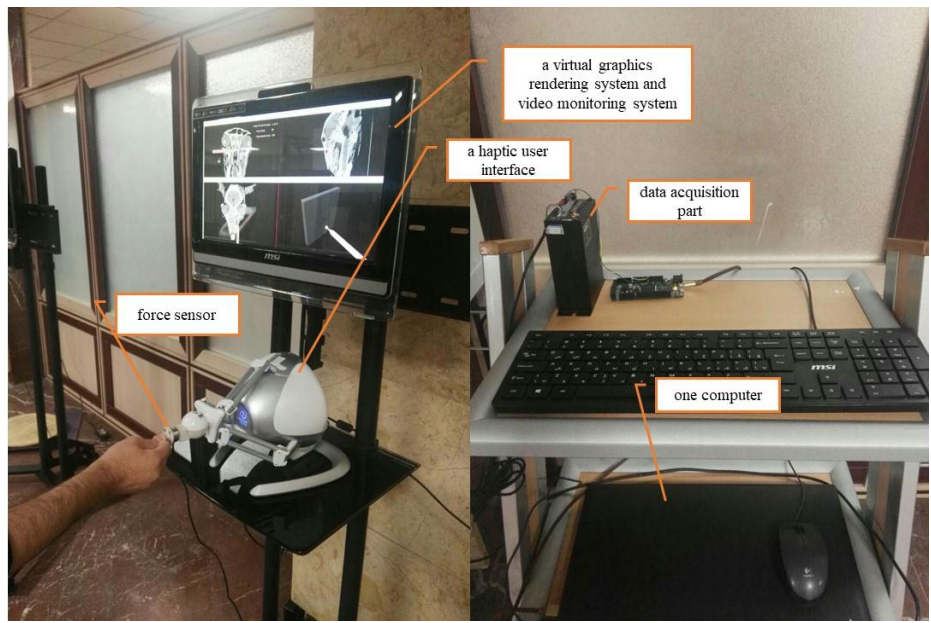

**Figure 3.3**. The endoscopic sinus surgery training system consists of a computer, virtual graphics rendering system, a haptic user interface, force sensor and data acquisition instruments, and a video monitoring system (Sadeghnejad *et al* 2019).

### 3.3.3 Anatomical Modelling

The tissue mechanics model computes the force-displacement relation of the coronal orbital floor tissue deformation and fracture. Considering the complexity of modern surgical techniques in the fields of otolaryngology and ophthalmology, the development of a model for virtual environment dynamics based on mechanical behaviour of tissue seems to be crucial.

The nonlinear behaviour of the sinus region caused by simultaneous existence of soft and hard tissues makes the tissue dynamic modelling challenging. Therefore, before simulating the interaction of human users with a haptic interface, we tried to select a proper dynamic model which realistically simulates the interaction phenomenon. One such model to be used in a training system of endoscopic sinus and skull base surgery that can also estimate the tissue fracture was suggested by Sadeghnejad et al (Sadeghnejad *et al* 2019) (Sadeghnejad *et al* 2016). A nonlinear rate-dependent model was proposed to predict the tissue behaviour prior and posterior to the fracture which is dependent on the tool's velocity and displacement while interacting with the tissue. The mechanical behaviour of the tissues is modelled as the measurement of stiffness, fracture, and cutting forces, and the effects of tool insertion rate during the penetration of simulated surgery tool into Sino-nasal tissues.

Another model, also proposed by Sadeghnejad et al., proposes hyperelastic modelling of Sino nasal tissue, specifically for haptic neurosurgery simulation. This mechanical modelling approach utilized optimization techniques with inverse finite element models to produce a model of hyperelastic behaviour of Sino nasal tissue. Through indentation experiments on sheep skulls, their force displacement curves proved to best fit a Yeoh hyperelastic model best, given its simplicity and low number of parameters (Sadeghnejad *et al* 2020). For the purposes of the proposed ESS system, the phenomenological tissue fracture modelling was employed.

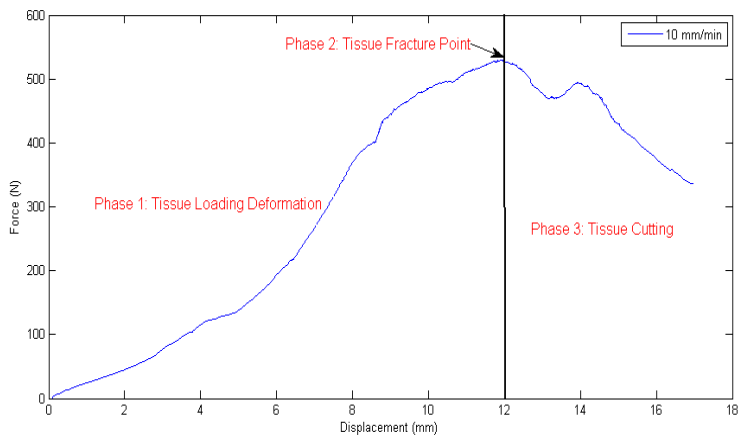

$$F(x, \dot{x}, t) = \begin{cases} F_1(x, \dot{x}) & x \leq X_f \\ F_2(x, \dot{x}) & x > X_f \end{cases}, \quad X_f(t) = H(\dot{x})$$

$$F_1 = F_s(x) + K(x)v\tau_s \left(1 - exp\left(-\frac{x}{v\tau_s}\right)\right)$$

$$F_2 = F_f(v) + a(v) * (x_{post} - x_f(v))$$

$$x_f(v) = 0.0001v^2 - 0.0575v + 19.21$$

$$F_s(x) = 0.008x^3 + 2.087x^2 + 8.766x$$

$$F_f(v) = 0.001v^2 - 1.176v + 697.1$$

$$a(v) = 10^{-7}v^4 + -7 \times 10^{-5}v^3 + 0.0101v^2 + 0.0485v + -79.313$$

**Figure 3.4**. The proposed nonlinear rate-dependent model to predict the tissue behaviour during the fracture (Sadeghnejad *et al* 2019)

### 3.3.4 Simulation

CHAI3D was used as the platform of choice for developing the virtual environment of the proposed ESS training simulator system. CHAI3D is a cross-platform C++ based simulation framework that supports a multitude of commercially available haptic devices, giving researchers the freedom to merge the visual and haptic feedbacks, combined with virtual Sino nasal tissues with realistic aesthetic and physical properties, thus allowing for a more realistic surgical simulation. CHAI3D supports several libraries to connect some haptic devices, such as Novint Falcon, Phantom, Delta, Omega, making it a versatile and adaptable tool.

Two main tasks were chosen for improving the performance of the trainees, which mimic the stepwise approaches employed in sinus surgery wall removals (which also represents the increasing level of the addressed difficulties). Surgically, the simulated environment can be seen clearly during training, and the relevant force haptic feedback can be predicted by the proposed tissue mechanical models, parallel to what may be sensed in a real operating room.

### 3.3.5  Haptics

Haptic feedback allows a user to employ one of the most critical senses during their training with a simulated training system; the sense of touch (Kolbari *et al* 2016) (Esfandiari *et al* 2017). Be it tactile or kinesthetic, accurate haptic feedback (i.e., feedback corresponding to realistic interaction between surfaces/objects) gives the trainee a better sense of the workspace, thus enabling them to potentially improve the quality and/or speed of procedure completion (Van Der Putten *et al* 2008) (Kolbari *et al* 2018) (Patel *et al* 2022).

One potential resource can be the Novint Falcon haptic device (Martin *et al* 2009). Falcon is a parallel impedance-type robot which is a robot of low price, considerable load capacities and proper workspace. The Falcon's programmable interface relieves the user from the inverse kinematics of the robot, which provides convenient control of the robot's motion on the three-motion axis (x, y, z). Due to its sampling frequency (1KH) and smooth actuation, operators use it in precise position sensing and high-fidelity motion control system (Martin *et al* 2009). Another potential haptic device is the Phantom Omni, a serial link based 6 D.O.F haptic device with a substantial workspace and producing a force up to 3.3 N. With its stylus like end-effector, users can simulate the use of a surgical tool, and can expect to experience the appropriate haptic feedback, assuming an accurate feedback model (Khadivar *et al* 2020). It was employed for their ESS simulator and is considered a device on the lower end of the quality spectrum, given its relatively low price point. To design a model predictive control scheme for a VR-based haptic system, we need to dynamically identify the haptic interface, by use of a calibrated force sensor to produce the proper robot impedance (Khadivar *et al* 2020). This haptic device was used in the development of our platform. The control system runs on a PC platform, and the interface between the sensor and the computer is an ARDUINO chip (Figure 3.5).

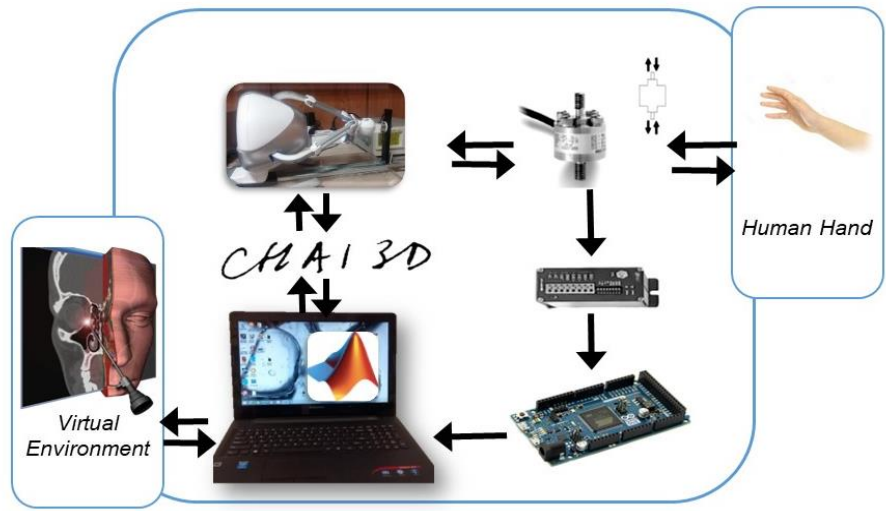

**Figure 3.5**. The whole components used in the development of a virtual-based haptic training system

The control system and the sensor update are implemented in CHAI3D open software. It should be noted that the dynamics of the Sino-nasal virtual environment were characterized as a linear parametric variable (LPV) problem with input constraints. We further employed an online robust model predictive control for a VR-based haptic system, inspired by a modified model predictive feedback (MPC) algorithm for LPV systems based on a quasi–min-max algorithm (Figure 3.6).

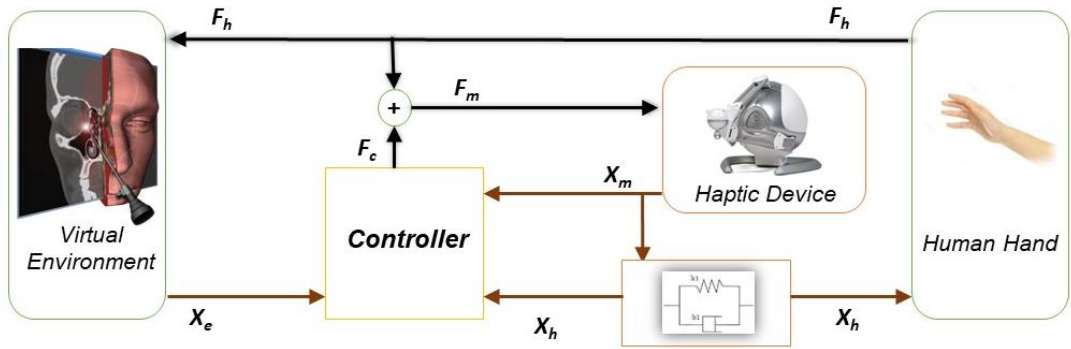

**Figure 3.6**. The schematic view components used in the design of the closed-loop control system

## 3.4 EXPERIMENTAL PROCEDURES AND EVALUATION

### 3.4.1 Experiment procedure

The system was tested on two teams of 10 members each, with one team receiving all the training through all the training tasks, and the second team getting limited exposure to training tasks. The participants were given the opportunity to partake in a pre-experiment learning session regarding the haptic interface and the virtual environment. Following this, the participants can then progress to participating in three specific tasks: **pre-training task**, **training task**, and the **evaluating task**. The pre-training task was aimed at familiarizing the participants with the system, including the haptic interface, the virtual reality environment, and the dynamics of tool-object interaction in the simulation. The training task involves using a curette (surgical tool) to indent simulated orbital floor tissue (which have variable stiffness which can be felt by the use via haptic feedback).

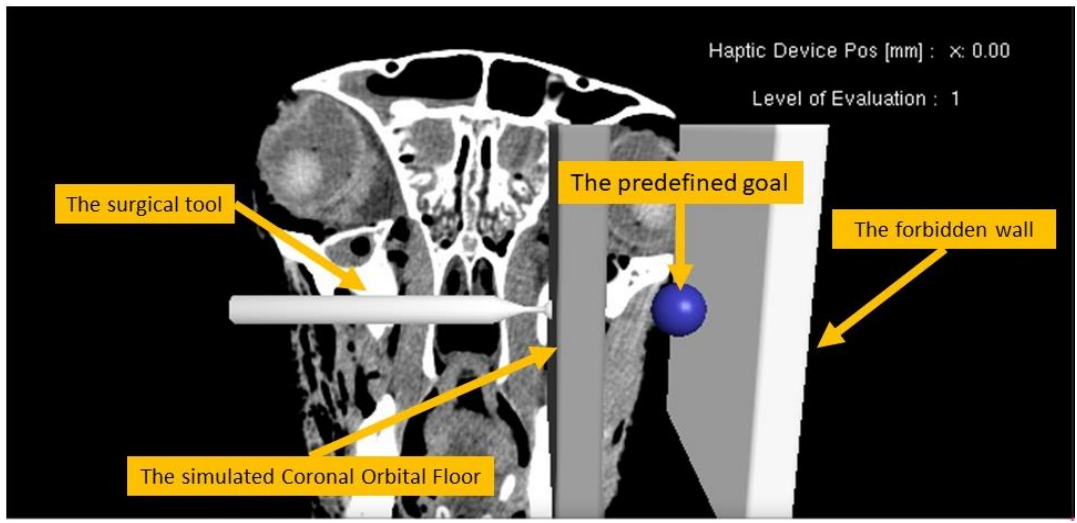

**Figure 3.7**. The evaluation task view: predefined goal, surgical tool, simulated Coronal Orbital Floor and forbidden wall.

Finally, the evaluating task aimed at improving the flexibility of the participants, by allowing them to navigate across a simulated orbital floor while avoiding virtual boundaries (Sadeghnejad *et al* 2019). The first group was allowed to partake in both the pre-training and training task before moving onto the evaluating task, while the second group was placed directly into the evaluating task. During the evaluation task, we recorded the generated interaction forces (both from force sensor and device's recording), the position of the surgery tool tip during the evaluation task, and the task completion time which is defined as the period when the tip of the surgery tool is in contact with the coronal orbital floor to the time of hitting the goal or the forbidden wall (Figure 3.7). The evaluating task has 5 different kinds of tissue stiffness, representing a spectrum of different performance levels. Participants were randomly assigned a level.

### 3.4.2 Performance metrics and evaluation methods

As discussed in previous sections, collecting, analysing, and reporting the user data is as crucial a step as any other in developing and proposing a new training simulator. Great care must be taken in measuring performance, by introducing metrics that can measure the performance of a trainee in an efficient, reproducible, actionable, and commensurable manner. Based on the task analysis process outlined above, our target dimensions of quantitative data generated covered the following areas: **quality** (Sense of haptic feedback of forces because of tool/anatomy interaction), **efficiency** (Task performance with the least number of unnecessary manoeuvres, as well as time taken), and **safety** (interaction with goal and the forbidden area).

The participants were also asked to fill out a questionnaire after the completion of the study, via a 10-point rating scale and open-ended questions, aimed at gathering qualitative data regarding their perceptions, opinions, and potential suggestions (Table 3.2).

**Table 3.2.** Description of the performance metrics

| Metric Sphere | Definition | Metric | Units |
|---|---|---|---|
| Quality | Sense of haptic feedback of forces, generated by tool-tissue interaction | • The generated force and the realism for orbital floor removal.<br>• How it will work for the surgical education curricula. | • Percentage<br><br>• Percentage |
| Efficiency | Task performance with the least number of unnecessary manoeuvres | • Distanced travelled to reach the goal post<br>• Time taken to reach the goal post | • Millimeters<br><br>• Seconds |
| Safety | Amount of simulated collateral damage | • How regular tool hits the simulated tissue as the goal<br>• How regular tool hits the simulated tissue normally not to be touched | • Number<br><br>• Number |

## 3.5 RESULTS

The performance of our developed system is examined by analysing the feedback of the participants in the Group I and Group II regarding the two criteria, namely, the sense of fracture and the sense of tissue stiffness variation. Based on the former criterion, the system was rated (out of 10) as $7.57 \pm 1.43$ and $6.32 \pm 1.02$ by the Group I and Group II, respectively, while the score associated to the system on the latter criterion was $8.13 \pm 1.87$ and $6.05 \pm 0.95$ for the two groups. Also, 77.1% of the participants of the Group I and 61.1% of those of the Group II expressed that they could adapt and improve their training level while working with the developed simulation system, which is substantial evidence that the system was appreciated by the users, in addition to the obvious difference in the performance metrics of the group with training vs the group without training. It should be noted that while 9/10 participants from the trained group reported their appreciation for using the developed simulation system as a supporting tool for hand eye coordination means, 7/10 members of the untrained group also reported the same belief, which only seeks to improve the credibility of the virtual-based haptic ESS training system (Table 3.3).

**Table 3.3**. Post-evaluation questionnaires assessment

| Evaluation Items[1] | Group I | | Group II | |
|---|---|---|---|---|
| | Mean Value | Standard Deviation | Mean Value | Standard Deviation |
| How users will sense the fracture during the operation | 7.57 | 1.43 | 6.32 | 1.02 |
| How users will sense the tissue hardening effect of tool interaction with the tissues | 8.13 | 1.87 | 6.05 | 0.95 |
| How effective will be the developed platform for education | 7.71 | 1.41 | 6.11 | 0.99 |

As it is clear from the result, there is a common considerable difference between two participants of each group in the educational programs. Although the participants of Group II did not recruit in the training tasks, the total time of evaluation task completion was reported to be much longer than the time taken by those who participated in Group I (Figure 3.8). It was also evident that those with the greater level of training had superior control over the trajectory of the path towards the specific goal, not to mention their ability to maintain steadier hand movements during the task, relative to the untrained group participants. It was also found that the untrained group, on average, exerted more force at the tooltip for the duration of the evaluation task (Figures 3.9 & 3.10).

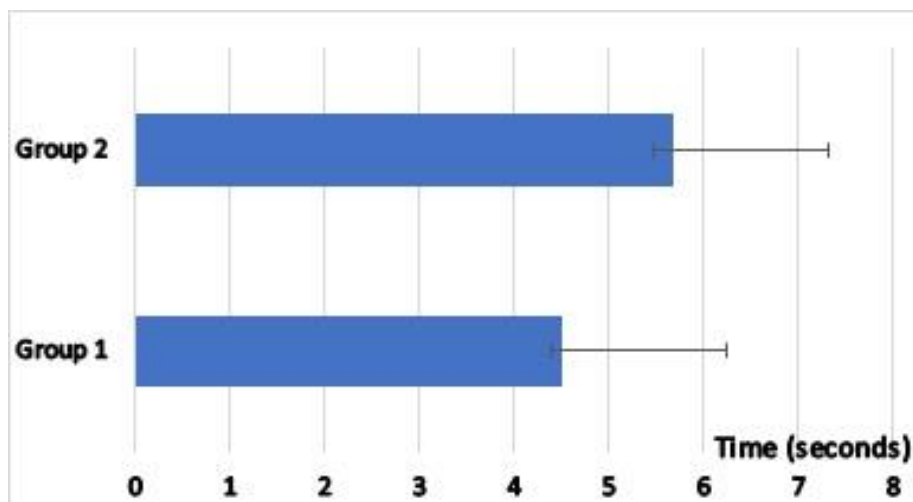

**Figure 3.8**. The average time used by in two different recruited groups for completing the evaluation task

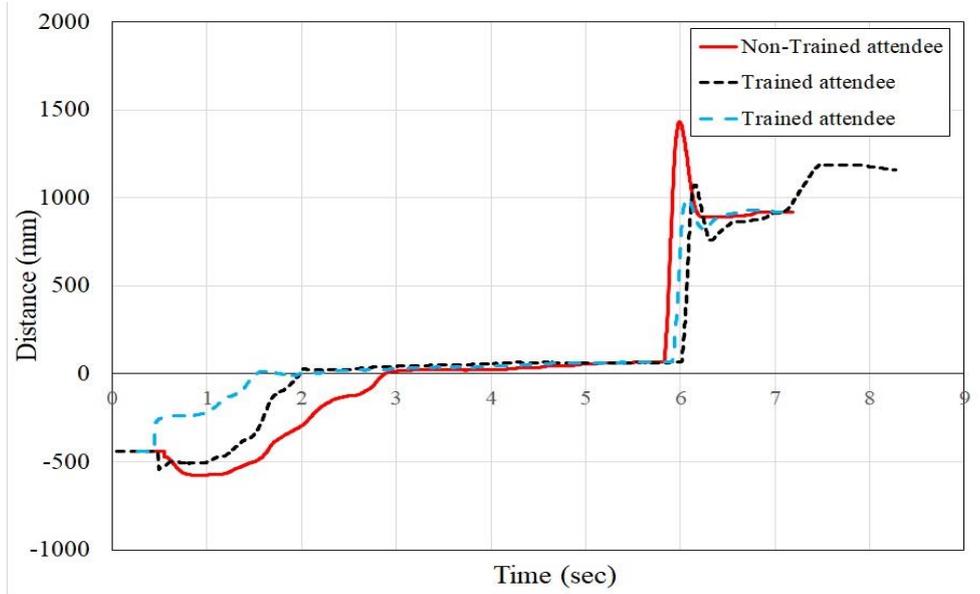

**Figure 3.9**. The average distance between tool tips through the simulation tasks

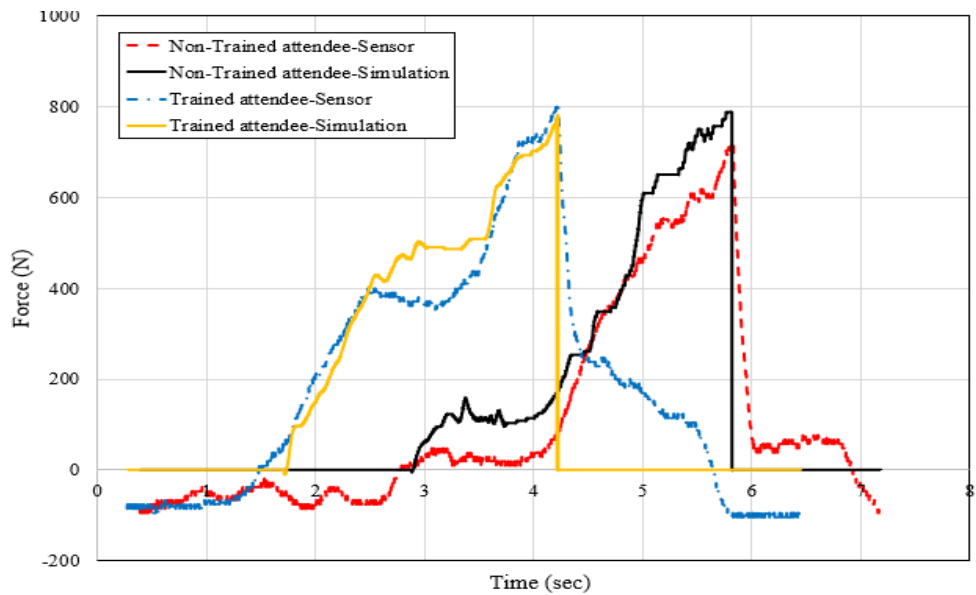

**Figure 3.10**. The average force of tool tips through the simulation tasks

The assessment of "safety" was conducted by comparing the number of times the predefined goal, after fracturing through the wall simulating the coronal orbital floor, was touched or by the number of times where the forbidden wall was hit. It is seen that most of the participants were generally able to show an efficient hand-eye coordination prior to the fracture, while maintaining such control over the tool after the fracture proved to be challenging as some participants hit both the predefined goal and the forbidden wall at the same time. 70% of the participants in the Group I and 50% of them in the Group II have been able to accomplish the evaluation task as expected, 25% in Group I and 35% in Group II hit the forbidden wall without touching the predefined goal, 5% in Group I and 15% in Group II touched both the predefined goal as well as the forbidden wall at the same time. The results of the safety assessment test demonstrate a significant effect of the training tasks over the performance of the participants working with the system (Figure 3.11).

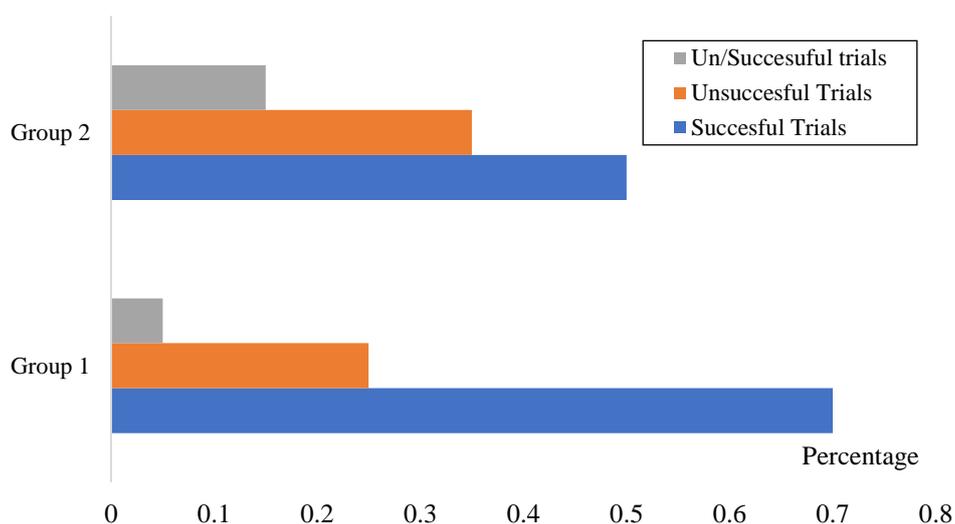

**Figure 3.11**. The successful trials percentage of simulated evaluation tasks in reaching to the specified goal through

## 3.6 CONCLUSION AND FINAL REMARKS

In this research, we developed an educational VR system with haptic force feedback from the interaction between the virtual tool and a graphical model of sinus tissue. To evaluate the system performance, we defined three different scenarios that requires doing specific tasks by the users. In the first phase, the pre-training phase, the users were asked to work with the system to get used to the virtual environment and the haptic interface. In the second phase, the training phase, the users were asked to exert higher force until fracture of Sino-nasal tissue takes place to perceive the fracture force level. Finally, we assessed the users' training levels and performance over the defined scenarios. The virtual environment is simulated in Chai3D open software. Using the Novint Falcon robot as the haptic interface, we provided haptic feedback to the user hand from the tool-tissue interaction force.

From the hardware setup to the anatomical model, to the test study specifics, we propose a full-scale approach to developing a EES training system that not only offers haptic feedback based on realistic tissue modelling (phenomenological tissue fracture modelling), but also delivers a variable testing protocol, which involves varying the level of task difficulties in different steps. Thus, the user, or trainee, is exposed to a more realistic experience with a more reliable physical and visual environment. Moreover, studying different performance metrics enables the researchers to develop a more comprehensive, and conclusive, assessment of the users' skills.

We proposed the evaluation of participants to be based on three distinct principles: quality, efficiency, and safety. This, in conjunction with a post-evaluation questionnaire test, has set precedence to enhance the system qualitatively and quantitatively. Based on the results reported, it is possible to develop an educational environment with the ability to simulate the haptic force resulting from the development of graphical simulation of tissue in the virtual environment. Based on average completion time alone in the evaluating test, it was evident that practice through the pre-training and training had noticeable positive effects on the performance of the participants. This point is further strengthened by the analysis of the applied force, the steadiness of the operating hand during the tasks, and overall better control over applied force. It was observed that the group with no access to the two training modules consistently applied a greater magnitude of force at tissue fracture compared to the group who had been trained through the pre-training and training task.

It is thus obvious that a surgeon can only benefit from pre-operative training, especially one that is conducted in a standardized environment which can be replicated for other trainees and trainers. In fact, this system reduces the need for a trainer, which can help make hospital operations more economical and efficient, while also reducing the overall risk of surgery by introducing a standardized method for training and evaluating any number of trainees.

Overall, ESS training systems with accurate tissue modelling and appropriate haptic feedback have a vast potential for shifting the current paradigm of cadaver and observation-based training methodologies, to a more sustainable, repeatable, economically viable, and adaptable training system.

## 3.7 CONFLICT OF INTEREST STATEMENT

The authors confirm that this book chapter and its corresponding research work involves no conflict of interest.

## 3.8 ACKNOWLEDGMENT

We are grateful to the Djavad Mowafaghian Research Center of Intelligent NeoruRehabilitation Technologies and the Research Center of Biomedical Technology and Robotics at the Research Institute of Medical Technology of Tehran University of Medical Sciences for the support of this research and conducting the experiments.